\documentclass[sn-basic]{sn-jnl}% Basic Springer Nature Reference Style/Chemistry Reference Style

\usepackage{graphicx}%
\usepackage{multirow}%
\usepackage{amsmath,amssymb,amsfonts}%
\usepackage{amsthm}%
\usepackage{mathrsfs}%
\usepackage[title]{appendix}%
\usepackage{xcolor}%
\usepackage{textcomp}%
\usepackage{manyfoot}%
\usepackage{makecell}
\usepackage{booktabs}
\usepackage{booktabs}%
\usepackage{algorithm}%
\usepackage{tabularx}
\usepackage{algorithmicx}%
\usepackage{algpseudocode}%
\usepackage{listings}%
\newcommand{\meanstd}[2]{\makecell[c]{$#1$\\[-0.2ex]{ $(#2)$}}}
\newcommand{\bestmeanstd}[2]{\makecell[c]{$\mathbf{#1}$\\[-0.2ex]{$(#2)$}}}
\usepackage{tablefootnote}
\theoremstyle{thmstyleone}%
\theoremstyle{thmstyletwo}%

\theoremstyle{thmstylethree}%

\begin{document}

\title[(G)Enformer]{Deep Generative Transformers for Probabilistic Time Series and Spatiotemporal Forecasting}


\author[1,2]{\fnm{Rajdeep} \sur{Pathak}}
\equalcont{These authors contributed equally to this work.}
\author[1,3]{\fnm{Rahul} \sur{Goswami}}
\equalcont{These authors contributed equally to this work.}
\author[1]{\fnm{Madhurima} \sur{Panja}}
\author[3]{\fnm{Palash} \sur{Ghosh}}
\author[1,2]{\fnm{Tanujit} \sur{Chakraborty}}\email{tanujit.chakraborty@sorbonne.ae}
\affil[1]{\orgname{SAFIR, Sorbonne University Abu Dhabi}, \country{United Arab Emirates}}
\affil[2]{\orgdiv{SCAI}, \orgname{Sorbonne Universit\'{e}}, \state{Paris}, \country{France}}
\affil[3]{\orgname{Indian Institute of Technology Guwahati}, \state{Assam}, \country{India}}


\abstract{Reliable uncertainty quantification is paramount for forecasting multivariate time series and spatiotemporal data. While Transformer architectures excel at sequence modeling, current probabilistic approaches typically rely on restrictive parametric likelihoods or quantile-based objectives, thereby limiting their ability to capture complex joint distributions in correlated time series. To overcome these limitations, we propose \textit{Enformer} and its spatiotemporal extension, \textit{GEnformer}. These models synthesize the expressive power of Transformers with engression, a stochastic learning paradigm for modeling conditional distributions. By injecting stochastic noise and optimizing a strictly proper scoring objective, our frameworks directly learn conditional predictive distributions without imposing parametric assumptions. This design ensures the generation of coherent multivariate trajectories while maintaining the Transformer's efficacy in modeling long-range dependencies and cross-series interactions. The probabilistic capability of Enformer is achieved with an asymptotic overhead of only a constant factor over a deterministic Transformer with an identical configuration. We extensively evaluate our frameworks on prominent multivariate benchmarks for temporal dynamics and real-world epidemic datasets for spatiotemporal dynamics. Empirical results demonstrate that both frameworks yield calibrated probabilistic forecasts and consistently outperform state-of-the-art baselines. Code is available at \url{https://github.com/yuvrajiro/genformer} and through our Python package \href{https://pypi.org/project/genformer/}{\texttt{genformer}}.}




\maketitle

\section{Introduction}
Time series forecasting plays a central role in many scientific and industrial applications, including energy management, traffic monitoring, financial analysis, and environmental modeling. Accurate forecasts enable informed decision-making in complex dynamical systems characterized by inherent uncertainty and intricate temporal and spatial dependencies. In many practical settings, forecasting involves multiple interacting time series, whose joint dynamics and cross-series or spatial dependencies must be modeled simultaneously to achieve reliable predictions. %\citep{yamazono2024permutation}.

% Time series forecasting plays a central role in many scientific and industrial applications, including energy management, traffic monitoring, financial analysis, and environmental modeling. Accurate forecasts enable informed decision-making in complex dynamical systems characterized by inherent uncertainty and intricate temporal and spatial dependencies. In many real-world settings, forecasting involves multiple interacting time series whose dynamics are influenced not only by their temporal evolution but also by cross-series and spatial dependencies. Effectively modeling these interactions is therefore essential for obtaining reliable forecasts.

Traditional statistical forecasting approaches, such as autoregressive models and state-space models (SSMs), have provided principled tools for modeling temporal dependencies. However, their scalability and expressiveness often deteriorate when applied to high-dimensional multivariate data. Recent advances in deep learning have therefore shifted attention towards neural sequence models, including recurrent neural networks (RNNs) \citep{hochreiter1997long} and, more recently, Transformer architectures \citep{vaswani2017attention}, which leverage self-attention mechanisms to capture long-range dependencies in sequential data. Transformers process sequences in parallel and model interactions between time steps directly through attention mechanisms, making them particularly effective for complex temporal patterns.

% Despite their success in deterministic sequence modeling, adapting Transformers to probabilistic forecasting remains challenging. Uncertainty quantification has become increasingly important for downstream tasks such as risk management, anomaly detection, and decision-making under uncertainty. Producing only point forecasts fails to capture the inherent variability of real-world systems and may obscure important risk information. Consequently, many existing deep probabilistic forecasting models rely on restrictive parametric likelihood assumptions or require carefully designed generative architectures to model predictive distributions. Probabilistic forecasting methods that integrate copula processes with RNNs \citep{salinas2019high}, an autoregressive deep learning with normalizing flows \citep{rasul2021multivariate}, diffusion models with guided learning process \citep{fan2024mgtsd}, or SSMs with Transformer architectures \citep{tang2021probabilistic} have been proposed to capture complex dependencies and generate predictive samples. While these approaches provide expressive uncertainty modeling, they often introduce additional architectural complexity or computational overhead.

Despite their success in deterministic forecasting, adapting Transformers for probabilistic forecasting remains challenging. In many forecasting applications, uncertainty quantification is pivotal as downstream tasks such as risk management, anomaly detection, and decision-making require an understanding of the range of plausible future outcomes. Point forecasts alone fail to capture the inherent variability of real-world systems and may therefore obscure important risk information. To address this limitation, numerous deep probabilistic forecasting approaches have been proposed. These include methods that combine copula processes with RNNs \citep{salinas2019high}, autoregressive deep learning models with normalizing flows \citep{rasul2021multivariate}, diffusion-based forecasting frameworks \citep{fan2024mgtsd}, and hybrid architectures integrating SSMs with Transformers \citep{tang2021probabilistic}. Although these methods offer expressive uncertainty modeling capabilities, they often rely on restrictive likelihood assumptions, specialized generative architectures, and computationally intensive training procedures.

Recent research has emphasized generative forecasting approaches that learn conditional predictive distributions and produce realistic trajectories rather than marginal intervals for spatiotemporal data \citep{kraft2026modeling, pathak2026deep}. %A promising direction for generative uncertainty modeling is distributional regression-based engression \citep{shen2025engression}, a framework that learns conditional distributions by injecting stochastic noise into model inputs and optimizing a strictly proper scoring objective. 
Within this emerging paradigm, distributional regression-based engression \citep{shen2025engression} has attracted considerable interest as a flexible framework for uncertainty modeling. Engression learns conditional distributions by injecting stochastic noise into model inputs and optimizing a strictly proper scoring objective. Unlike many generative approaches that require specialized architectures or complex training procedures, engression can be integrated with existing neural models with minimal modification while enabling sample-based uncertainty quantification. 

Motivated by these developments, we propose \textit{Enformer} and \textit{Graph-Enformer (GEnformer)} - two deep generative forecasting frameworks that integrate the engression principle with Transformer architectures for probabilistic temporal and spatiotemporal forecasting. The proposed models leverage the attention mechanism of Transformers to capture long-range temporal and cross-series dependencies while employing noise-driven training to learn the full conditional predictive distribution. For spatiotemporal forecasting, GEnformer further incorporates graph convolution operations to model complex spatial interactions among interconnected locations. This combination enables the generation of diverse and realistic forecast trajectories while quantifying predictive uncertainty without imposing restrictive distributional assumptions or requiring substantial architectural modifications. Furthermore, a complexity analysis shows that the probabilistic capability of Enformer is obtained with only an asymptotic constant-factor overhead relative to a deterministic Transformer of identical configuration, thereby preserving the favorable scalability of the underlying architecture.

% Motivated by these developments, we propose \textit{Enformer} and \textit{Graph-Enformer} \textit{(GEnformer)} - deep generative frameworks integrating the engression principle with Transformer architectures, for probabilistic temporal and spatiotemporal forecasting. The proposed approaches leverage the attention mechanism of Transformers to model long-range temporal and cross-series dependencies while employing noise-driven training to learn the full predictive distribution. GEnformer utilizes graph convolution operations in its spatial module to capture complex spatial dynamics. Such combination enables the models to generate diverse plausible forecast trajectories and quantify uncertainty without imposing restrictive distributional assumptions or requiring substantial architectural changes. A complexity analysis reveals that the probabilistic capability of Enformer is obtained at an asymptotic overhead of a constant factor $M$ over a deterministic Transformer of identical configuration, preserving the favorable scaling of the underlying architecture. 
%Moreover, it facilitates uncertainty quantification while remaining computationally lightweight and scalable for complex multivariate time series systems. 

To evaluate the proposed frameworks, we conduct experiments on several widely used benchmarks for multivariate probabilistic forecasting, including electricity demand, traffic flow, solar power generation, and other real-world time series. Given the importance of uncertainty quantification in epidemic forecasting, GEnformer is further evaluated on five spatiotemporal epidemic datasets with diverse transmission dynamics and geographical structures. Across these benchmarks, the proposed models demonstrate strong predictive performance and well-calibrated probabilistic forecasts. The main contributions of this work are summarized as follows:
% We assess Enformer using several widely adopted benchmark datasets for multivariate probabilistic forecasting, including electricity demand, traffic flow, solar power generation, and other real-world time series. As epidemic forecasting is a critical task requiring robust uncertainty quantification, GEnformer is evaluated on 5 spatiotemporal epidemic datasets of varying transmission modes and geographical topology. The empirical results show that the proposed frameworks deliver strong predictive accuracy while providing well-calibrated probabilistic forecasts. The main contributions of this study are summarized as follows:
\begin{enumerate}
    \item We propose Enformer, a lightweight Transformer-based framework designed for accurate probabilistic multivariate time series forecasting, and its spatiotemporal variant, GEnformer, which integrates graph convolutional networks (GCNs) with Enformer for uncertainty-aware spatiotemporal forecasting. 
    % We show how engression can be integrated with Transformer architectures to learn predictive distributions and generate realistic forecast trajectories.
    \item We demonstrate how engression can be incorporated into Transformer architectures to learn predictive distributions and generate realistic forecast trajectories.
    % \item The probabilistic functionality of the Enformer incurs merely a linear asymptotic overhead over its deterministic counterpart, thereby retaining the core architectural scalability of the canonical Transformer.
    \item We show that the probabilistic capability of Enformer is achieved with only a linear asymptotic overhead relative to its deterministic counterpart, preserving scalability.
    % \item The effectiveness of the proposed method is demonstrated through empirical evaluation on multiple real-world forecasting benchmarks.
    \item We validate the effectiveness of the proposed frameworks through empirical evaluation on multiple real-world temporal and spatiotemporal forecasting benchmarks.
\end{enumerate}

The rest of this paper is organized as follows. Sec. \ref{sec:bg} outlines the problem formulation and reviews Transformer-based approaches for time series forecasting. Sec. \ref{sec:proposed_frameworks} introduces the proposed frameworks. Sec. \ref{sec:experimental_setup} details the experimental setup and presents the empirical evaluation on benchmark datasets. Finally, Sec. \ref{sec:conclusion} summarizes the main findings and highlights possible directions for future research.

\section{Background} \label{sec:bg}
\subsection{Problem Formulation and Notations} \label{sec:problem_formulation}
Let $\boldsymbol{Y} \in \mathbb{R}^{T \times D}$ denote a multivariate time series dataset, where $T$ is the total number of observed time steps and $D$ indicates the feature dimension. In a spatiotemporal context, $D$ denotes the number of spatial locations or nodes. Let ${\boldsymbol{y}}_{t} \in \mathbb{R}^D$ represent the observations for all $D$ nodes of $\boldsymbol{Y}$ at time step $t$. Consider a historical (look-back) window of length $p$, denoted by $\boldsymbol{Y}_{t-p+1:t}=\left[\boldsymbol{y}_{t-p+1}, \ldots, \boldsymbol{y}_{t}\right]$. The objective of the time series forecasting problem is to estimate the future sequence over a prediction horizon $q$, expressed as $\widetilde{\boldsymbol{Y}}_{t+1: t+q}=\left[\widetilde{\boldsymbol{y}}_{t+1}, \ldots, \widetilde{\boldsymbol{y}}_{t+q}\right]$, where $\widetilde{\boldsymbol{y}}_\tau \in \mathbb{R}^D$ denotes the forecast at time step $\tau$.
To accomplish this, a forecasting model $f\left(\boldsymbol{Y}_{t-p+1: t}, \Theta\right)$ (for instance, a Transformer-based network) is learned to map the past observations to future values over the horizon $q$.

While the formulation above describes point forecasting, many real-world applications require reliable quantification of predictive uncertainty. In such settings, the objective is not merely to predict a single trajectory, but to learn the conditional distribution of future observations given past data. Formally, this can be written as
$$
\widetilde{\boldsymbol{Y}}_{t+1: t+q} \sim p_{\widetilde{\Theta}}\left(\cdot \mid \boldsymbol{Y}_{t-p+1: t}\right),
$$
where $p_{\widetilde{\Theta}}$ denotes the predictive distribution parameterized by $\widetilde{\Theta}$. Although Transformer architectures have demonstrated remarkable success in modeling long-range dependencies in time series, an important question arises: {\it How can Transformers be effectively adapted to generate coherent probabilistic forecasts rather than point predictions?} 

Existing approaches typically rely on parametric likelihood assumptions \citep{salinas2020deepar, rasul2021autoregressive}, quantile-based objectives \citep{lim2021temporal}, or computationally heavy generative models \citep{tang2021probabilistic, fan2024mgtsd} which may restrict the flexibility of the learned predictive distribution in high-dimensional multivariate and spatiotemporal settings. To address this limitation, we adopt the engression framework \citep{shen2025engression}, a stochastic learning paradigm that enables neural networks to learn conditional distributions via noise-driven sampling. This idea enables Transformers to produce flexible, uncertainty-aware probabilistic forecasts.

%By injecting stochastic noise into the forecasting model and training it with a proper scoring objective, engression enables the model to produce diverse `plausible' forecast trajectories that approximate the conditional distribution of future values. 

\subsection{Transformer Architectures for Time Series Forecasting}
The core of the Transformer architecture \citep{vaswani2017attention} is the self-attention mechanism, which enables dynamic modeling of dependencies across sequence positions. In time series forecasting, this allows the model to capture temporal relationships within a historical lookback window $\boldsymbol{Y}_{t-p+1:t}$. Given queries $Q \in \mathbb{R}^{\ell_q \times d}$ and corresponding keys and values $K, V \in \mathbb{R}^{\ell_k \times d}$, multi-head attention computes an output sequence $O = [O_1, \ldots, O_H]$ across $H$ independent heads:$$O_h = \operatorname{Attention}(Q_h, K_h, V_h) = \operatorname{Softmax}\left(\frac{Q_h K_h^{\top}}{\sqrt{d}}\right) V_h,$$where $Q_h = Q W_h^Q, K_h = K W_h^K$, and $V_h = V W_h^V$ are linear projections governed by learnable weight matrices. Self-attention arises when $Q = K = V$. Because attention operations are inherently permutation equivariant, temporal ordering is preserved by injecting sinusoidal positional encodings $p_t(i)$ into the attention mechanism:
$$p_t(i)= \begin{cases} \sin \left(t \cdot c^{i / d}\right), &i \text{ even}\\ \cos \left(t \cdot c^{i / d}\right), &i \text{ odd} \end{cases},$$
where $c$ is a scaling constant. While numerous Transformer variants have been developed to capture long-range dependencies in multivariate time series (e.g., Autoformer \citep{wu2021autoformer}) and spatiotemporal domains (e.g., STFT \citep{wang2024spatiotemporal}), these architectures either primarily yield deterministic point forecasts, or are extremely computationally expensive (e.g., ProTran \citep{tang2021probabilistic}). To bridge these gaps, we propose computationally lightweight Transformer frameworks designed specifically for probabilistic forecasting. Our approach requires minimal architectural modifications to existing training procedures while naturally enabling model-intrinsic uncertainty quantification.

%\subsection{Engression}

\section{Proposed Methodology} \label{sec:proposed_frameworks}
In this section, we detail the architectures and training mechanisms of Enformer and GEnformer. Building upon the notations established in Sec. \ref{sec:problem_formulation}, we describe how the proposed models are formulated for probabilistic forecasting by incorporating a pre-additive noise model into the Transformer architecture while ignoring any restrictive parametric assumptions. We also outline how Enformer integrates multivariate auxiliary features and optimizes a strictly proper scoring rule, and provide a brief discussion on its computational complexity. 

% The proposed models produce probabilistic forecasts using the engression principle, which is a deep distributional regression methodology with pre-additive noise mechanism. Originally introduced in a regression setup \citep{shen2025engression}, engression has demonstrated superior extrapolation capabilities relative to conventional neural networks and tree-based algorithms. Subsequently, this framework has been adapted for more complex sequential and spatial domains, including time series rainfall-runoff modeling \citep{kraft2026modeling} and the probabilistic forecasting of epidemic dynamics across space and time \citep{pathak2026deep}.

\subsection{The Enformer Architecture}
Multivariate forecasting tasks frequently require models to process auxiliary features alongside the primary target time series. To accommodate this, Enformer is structured to natively ingest past covariates $\boldsymbol{C}\in \mathbb{R}^{T\times D_{cov}}$, where $D_{cov}$ denotes the number of auxiliary variables. Operating as a sequence-to-sequence architecture, the proposed model processes a historical look-back window of length $p$ at each time step $t$ to forecast the subsequent $q$ steps. Specifically, the historical target sequence $\boldsymbol{Y}_{t-p+1:t}$ and the corresponding available past covariates $\boldsymbol{C}_{t-p+1:t}$ are concatenated along the feature dimension. This yields a consolidated input matrix $\boldsymbol{X} \in \mathbb{R}^{p \times (D + D_{cov})}$, such that $\boldsymbol{X}_{t-p+1:t} = [\boldsymbol{Y}_{t-p+1:t} \parallel \boldsymbol{C}_{t-p+1:t}]$. To enhance computational efficiency, the input sequences and covariates from the training data are structured into mini-batches, yielding an effective input tensor denoted by $\boldsymbol{X}_{batch} = \left\{\boldsymbol{X}_{t-p+1:t};\ p\leq t \leq T-q\right\} \in \mathbb{R}^{B \times p \times (D+D_{cov})}$, where $B$ denotes the batch size. 

\begin{figure}[!t]
\centering
\includegraphics[width=0.9\textwidth]{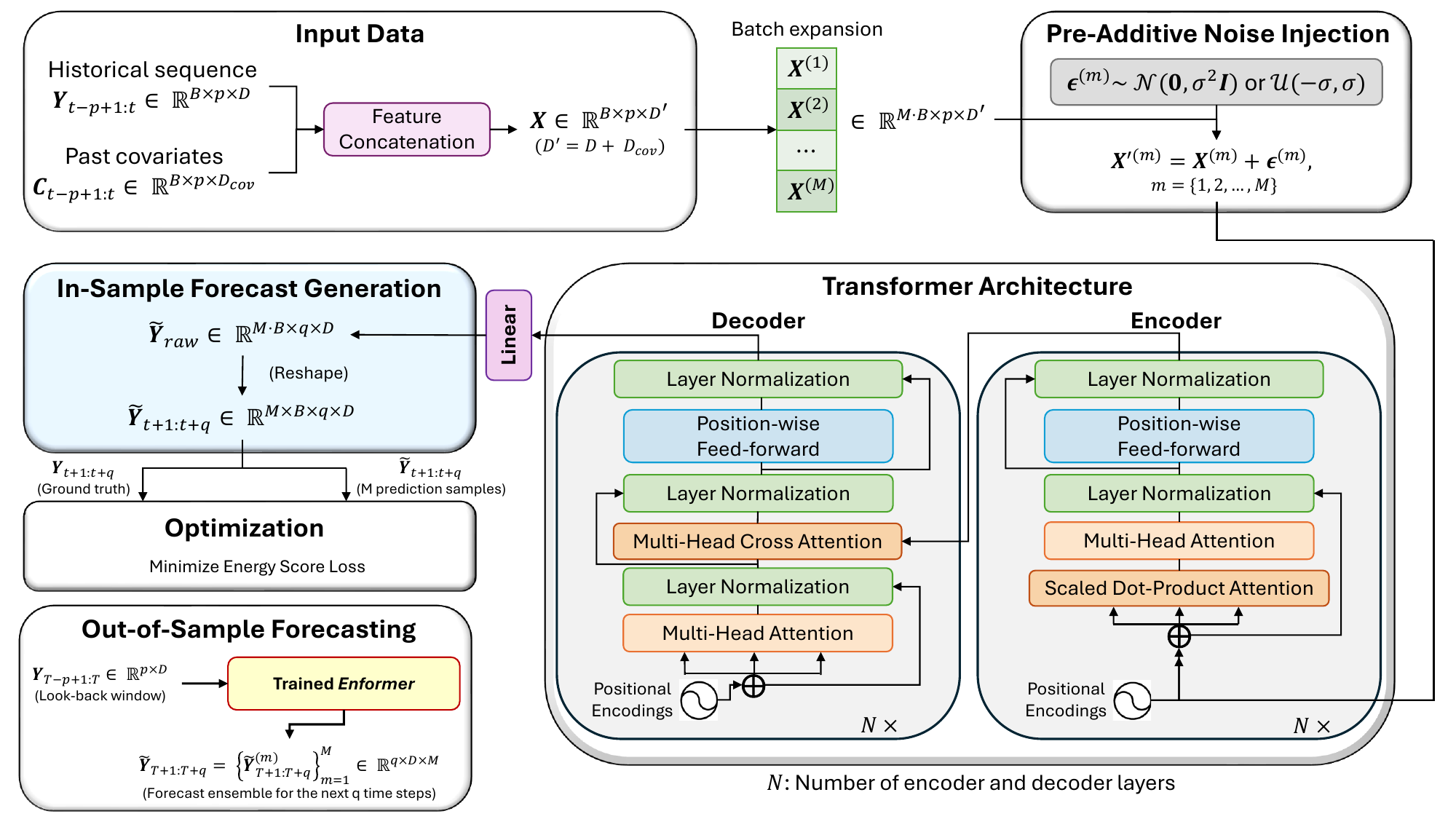}
\caption{Overview of the Enformer architecture, consisting of noise injection, in-sample forecast generation, and optimization using the energy score loss. An ensemble of out-of-sample forecasts for all $D$ nodes for the next $q$ steps can be generated from the trained model by passing the $p$-length look-back window as input.} \label{fig:entransformer_architecture}
\end{figure}

The core element of the Enformer architecture is the pre-additive stochastic noise injection, which enables the generation of probabilistic forecasts. The input matrix is explicitly duplicated $M$ times along the batch dimension (RepeatInterleave operation), yielding a batched tensor of shape $(M \times B, p, D + D_{cov})$. A stochastic noise layer is appended directly to the Transformer encoder input. For each of the $M$ duplicated sequences $\boldsymbol{X}^{(m)}_{batch}$, an independent noise tensor $\boldsymbol{\epsilon}^{(m)}$ of the same dimension is sampled and added to the raw features before they are linearly projected into the model's hidden dimension, as $\boldsymbol{X'}_{batch}^{(m)} = \boldsymbol{X}_{batch}^{(m)} + \boldsymbol{\epsilon}^{(m)}.$ 

The noise $\boldsymbol{\epsilon}^{(m)}$ is sampled from a Gaussian or Uniform distribution, i.e., $\boldsymbol{\epsilon}^{(m)} \sim \mathcal{N}(\boldsymbol{0}, \sigma^2 \mathbf{I})$ or $\boldsymbol{\epsilon}^{(m)} \sim \mathcal{U}(-\sigma, \sigma)$, where $\sigma$ is a predefined hyperparameter dictating the standard deviation of the injected noise. The choice of the noise distribution is data dependent and can be empirically determined through standard validation techniques. By expanding the input sequence $M$ times along the batch dimension prior to the encoder, the architecture requires only a single forward pass to generate the stochastic ensemble. During this pass, the sampled noise tensor matching the augmented batch shape simultaneously applies independent stochastic perturbations to each of the $M$ replicas. Consequently, the initially identical historical contexts are efficiently mapped to $M$ mathematically distinct continuous representations within a single computational step. 

These noise-perturbed sequences $\boldsymbol{X'}^{(m)}_{batch}$ are subsequently processed by the multi-head self-attention layers of the Transformer network. By propagating these stochastic representations through the network, the architecture generates a diverse set of $M$ plausible multi-step-ahead forecast trajectories $\{\widetilde{\boldsymbol{Y}}^{(m)}_{t+1:t+q}\}_{m=1}^M$ for the same horizon, that collectively approximate the complex, non-parametric conditional predictive distribution. The Enformer framework is generative, since different plausible forecast trajectories can be generated by perturbing the input sequences with different noise samples, effectively sampling from the predictive distribution. Consequently, this forecast ensemble facilitates the derivation of robust point forecasts (such as the empirical median) and allows the construction of probabilistic prediction intervals by extracting specific empirical quantiles. The pseudo-code for a forward pass of the Enformer architecture is presented in Algorithm \ref{alg:entransformer_forward}.

\subsubsection{Optimization}
Since Enformer is designed to directly learn the conditional predictive distribution without imposing parametric assumptions, standard maximum likelihood estimation techniques based on fixed parametric distributions are inapplicable. Instead, the proposed approach injects noise into the model representation and optimizes an energy-based scoring objective. The Energy Score is a strictly proper scoring rule \citep{gneiting2007strictly} that assesses the quality of multivariate probabilistic forecasts by directly operating on the empirical distribution of the generated samples. For a time step $t$, given the ground truth observed sequence $\boldsymbol{Y}_{t+1:t+q}$ and the set of $M$ generated in-sample forecast trajectories denoted by $\widetilde{\boldsymbol{Y}}^{(m)}_{t+1:t+q}$ for $m \in \{1, \dots, M\}$, the empirical energy score (ES) loss is computed as:
{\small
\begin{equation*}
%\begin{split}
\mathcal{L}_{ES} = \frac{1}{M} \sum_{m=1}^{M} \left\| \widetilde{\boldsymbol{Y}}^{(m)}_{t+1:t+q} - \boldsymbol{Y}_{t+1:t+q} \right\|_2  - \frac{1}{2 M(M-1)} \sum_{i=1}^{M} \sum_{j=1}^{M} \left\| \widetilde{\boldsymbol{Y}}^{(i)}_{t+1:t+q} - \widetilde{\boldsymbol{Y}}^{(j)}_{t+1:t+q} \right\|_2.
%\end{split}
\end{equation*}}
This loss function balances two competing geometric objectives for stochastic learning. The first term minimizes the expected Euclidean norm between the generated samples and the ground truth, ensuring the predictive distribution remains accurately centered. The second term acts as a regularizer by maximizing the expected pairwise Euclidean distance among all $M$ independent samples. This promotes trajectory dispersion and prevents the model from collapsing into a deterministic point forecaster. Consequently, minimizing this joint objective allows Enformer to generate sharp, well-calibrated, and adequately dispersed multivariate trajectories that capture true predictive uncertainty. Furthermore, the strict propriety of the loss function mathematically guarantees convergence to the true distribution for well-specified models. Algorithm \ref{alg:entransformer_training} outlines the end-to-end training process and Fig. \ref{fig:entransformer_architecture} illustrates the architecture.

\begin{algorithm}[H]
\caption{Enformer: Training Step and Energy Score Optimization}
\label{alg:entransformer_training}
\begin{algorithmic}[1]
\Require Batched historical target sequences: $\boldsymbol{Y}_{batch} \in \mathbb{R}^{B \times p \times D}$; Covariates batch: $\boldsymbol{C}_{batch} \in \mathbb{R}^{B \times p \times D_{cov}}$ (optional); Batched ground truth future target sequences: $\boldsymbol{Y}_{true} \in \mathbb{R}^{B \times q \times D}$; Number of in-sample forecast trajectories: $M$
\Ensure Energy Score Loss $\mathcal{L}_{ES}$

\Statex \textbf{// 1. Covariate integration and sequence expansion}
\State $\boldsymbol{X}_{batch} \leftarrow [\boldsymbol{Y}_{batch} \parallel \boldsymbol{C}_{batch}] \in \mathbb{R}^{B \times p \times D'}$ \Comment{$D' = D + D_{cov}$}

\State $\boldsymbol{X}_{batch}^{(M)} \leftarrow \text{RepeatInterleave}(\boldsymbol{X}_{batch}, M, \text{dim}=0) \in \mathbb{R}^{(M \cdot B) \times p \times D'}$ \Comment{Duplicate batch $M$ times}
\Statex \textbf{// 2. Forecast ensemble generation}

\State $\widetilde{\boldsymbol{Y}}_{raw} \leftarrow \text{ForwardPass}\left(\boldsymbol{X}_{batch}^{(M)}\right) \in \mathbb{R}^{(M \cdot B) \times q \times D}$ \Comment{Calls Algorithm \ref{alg:entransformer_forward}}
\State $\widetilde{\boldsymbol{Y}}_{pred} \leftarrow \text{View}\left(\widetilde{\boldsymbol{Y}}_{raw}, M, B, q, D\right)$ \Comment{Separate batch and sample dimensions to get in-sample forecast ensemble}
\Statex \textbf{// 3. Loss computation}
\State Loss $= \mathcal{L}_{ES}\left(\boldsymbol{Y}_{true}, \widetilde{\boldsymbol{Y}}_{pred}\right)$

\State \Return Loss
\end{algorithmic}
\end{algorithm}

\begin{algorithm}[H]
\caption{Enformer: Forward Pass}
\label{alg:entransformer_forward}
\begin{algorithmic}[1]
\Require Expanded input sequence batch $\boldsymbol{X}_{batch}^{(M)} \in \mathbb{R}^{MB \times p \times D'}$ ($D' = D + D_{cov}$);  Noise standard deviation: $\sigma$; Noise distribution (Gaussian or Uniform) 
\Ensure Forecast output $\widetilde{\boldsymbol{Y}}_{raw} \in \mathbb{R}^{MB \times q \times D}$

\Statex \textbf{// 1. Stochastic noise injection}
\If{NoiseDistribution == Gaussian}
    \State Sample $\boldsymbol{\epsilon} \sim \mathcal{N}(\boldsymbol{0}, \sigma^2 \boldsymbol{I})$ of shape $(MB, p, D')$
\ElsIf{NoiseDistribution == Uniform}
    \State Sample $\boldsymbol{\epsilon} \sim \mathcal{U}(-\sigma, \sigma)$ of shape $(MB, p, D')$
\EndIf
\State $\boldsymbol{X'}^{(M)}_{batch} \leftarrow \boldsymbol{X}^{(M)}_{batch} + \boldsymbol{\epsilon}$ \Comment{Perturb the input sequence}

\Statex \textbf{// 2. Transformer sequence processing}
\State $\boldsymbol{H}_{enc} \leftarrow \text{TransformerEncoder}\left(\boldsymbol{X'}^{(M)}_{batch}\right)$ \Comment{Process perturbed history}
\State $\widetilde{\boldsymbol{Y}}_{raw} \leftarrow \text{TransformerDecoder}(\boldsymbol{H}_{enc}) \in \mathbb{R}^{MB \times q \times D}$ \Comment{Decoding}
\State \Return $\widetilde{\boldsymbol{Y}}_{raw}$
\end{algorithmic}
\end{algorithm}

\subsubsection{Complexity Analysis} \label{sec:complexity_analysis}
The computational complexity of Enformer is governed by its $M$-fold stochastic batch expansion, which processes replicas simultaneously to ensure the multi-head attention backbone scales strictly linearly with the ensemble size $M$. Initializing the noise perturbation and linear projection requires $\mathcal{O}(MBpD')$ and $\mathcal{O}(MBpD'd)$ time, respectively (where $d$ is the hidden dimension of linear projection). Processing the look-back sequence $p$ and forecast horizon $q$ through the $L$-layer Transformer takes $\mathcal{O}(MB L(p^2 d + p d^2))$ for the encoder and $\mathcal{O}(MB L(q^2 d + pq d + q d^2))$ for the decoder \citep{keles2023computational}. The sole component exhibiting quadratic dependence on $M$ is the ES loss, costing $\mathcal{O}(B M^2 q D)$ due to its pairwise trajectory dispersion evaluation. Consequently, the total time complexity per training step is
$$
\mathcal{O}\Big(\underbrace{MB\,L\big(p^2 d + q^2 d + pq\,d + (p+q)d^2\big)}_{\text{Transformer backbone}} \;+\; \underbrace{MBpD'd}_{\text{projection}} \;+\; \underbrace{BM^2 qD}_{\text{energy score}}\Big).
$$

Spatially, the memory footprint is dominated by the $M$-expanded hidden activations and layer-wise attention maps, yielding a total memory complexity of $\mathcal{O}\big(MB L(p+q)^2 + MB(p+q)d\big)$. Two key operational benefits emerge: first, the model retains the standard Transformer's $\mathcal{O}((p+q)^2)$ quadratic scaling with respect to sequence length, unaffected by the noise injection. Second, the $\mathcal{O}(M^2)$ overhead is safely isolated to the ES loss, which is computationally cheap relative to attention whenever $M\,D \ll L(p+q)^2 d$. Moreover, the pairwise distances can be computed efficiently using vectorization on GPUs. The transition from a deterministic Transformer to a probabilistic framework thus imposes only a constant-factor asymptotic penalty, bounded by $M$, on the Enformer. As such, the model secures robust uncertainty quantification without compromising the fundamental efficiency and scaling properties of the underlying self-attention mechanism.

\subsection{GEnformer}
For spatiotemporal forecasting, we further introduce Graph-Enformer, which extends Enformer to jointly model temporal dynamics, spatial interactions, and predictive uncertainty. The framework incorporates a spatial module in which the topology of the locations is represented by a static adjacency matrix $\mathbf{A}\in\mathbb{R}^{D\times D}$, while spatial dependencies are captured through GCNs to obtain spatiotemporal embeddings. Specifically, $\mathbf{A}=[a_{u,v}]$ represents an undirected graph with $D$ vertices and is constructed using a thresholded Gaussian kernel applied to the Haversine distance $s_{u,v}$ between nodes $u$ and $v$ \citep{panja2026stgcn}. The edge weights are thus defined as: $$a_{u,v} = \exp\left(-\frac{s_{u,v}^2}{\rho}\right), \text{ when } u \neq v \text{ and } \exp\left(-\frac{s_{u,v}^2}{\rho}\right) \geq \tau,$$where $\rho$ and $\tau$ regulate spatial variance and matrix sparsity, respectively. Operating in a sequence-to-sequence paradigm, the module maps a $p$-step temporal slice $\boldsymbol{Y}_{t-p+1:t} \in \mathbb{R}^{p\times D}$ and topology $\mathbf{A}$ to a latent embedding $\mathbf{H}_{t-p+1:t} \in \mathbb{R}^{p\times \widetilde{D}}$:$$\mathbf{H}_{t-p+1:t} = \mathrm{GCN}(\boldsymbol{Y}_{t-p+1:t}, \mathbf{A}; \mathbf{\Omega}),$$with $\mathbf{\Omega}$ and $\widetilde{D}$ denoting the learnable GCN weights and dimension of the spatial embeddings, respectively. The learning procedure follows a neural message passing mechanism as detailed in \cite{panja2026stgcn, pathak2026deep}. The spatially aware embeddings are subsequently passed to the Enformer architecture for temporal processing and forecasting with uncertainty.

\section{Experiments} \label{sec:experimental_setup}
We perform extensive experiments to comprehensively assess the forecasting performance and generalization capability of the proposed frameworks. In the following sections, we describe the experimental setup and the forecasting performance of the proposed and state-of-the-art architectures.

% The empirical analysis for Enformer is conducted on six multivariate benchmark time series datasets representing diverse domains and temporal dynamics. To evaluate GEnformer, we utilize five real-world spatiotemporal epidemic datasets spanning distinct diseases and geographical structures.

\subsection{Experimental Setup}
\subsubsection{Datasets} 
We evaluate the performance of the Enformer model on six multivariate temporal datasets, namely Solar, Electricity, Traffic, KDD-cup, Taxi, and Wikipedia, covering diverse application domains \citep{alexandrov2020gluonts}. The Solar dataset provides synthetic US photovoltaic generation data for energy forecasting; Electricity details hourly power consumption across 370 distinct clients; Traffic records road occupancy rates (ranging from 0 to 1) from 963 highway sensors in the San Francisco Bay Area; KDD-cup contains multivariate air-quality measurements from monitoring stations in Beijing and London; the Taxi dataset comprises comprehensive NYC trip records for urban mobility and demand forecasting; and Wikipedia tracks temporal online user activity via daily article page views. These benchmark datasets are frequently utilized in the time series literature to evaluate state-of-the-art models. The global characteristics of these datasets and the source links are provided in Table~\ref{tab:temporal_data_characteristics}. For spatiotemporal forecasting, we consider 5 epidemic datasets available in the benchmark established by \cite{panja2026epicastbench} - tuberculosis (TB) in Japan, influenza-like illnesses (ILI) in the USA, COVID-19 in Belgium, dengue in Colombia, and chickenpox in Hungary. The properties of the epidemic datasets are listed in Table \ref{table:spatiotemporal_dataset_prop}.

\begin{table}[]
    \centering
    \caption{Key characteristics of the temporal benchmark datasets used in this study.}
\label{tab:temporal_data_characteristics}
    \begin{tabular}{cccccccc}
    \toprule Name & Freq. & Dim. & \makecell{Training\\Samples} & \makecell{Context\\Length} & \makecell{Pred.\\Length} & \makecell{Seq.\\Lags} & \makecell{Test\\ Windows}   \\
   
    \midrule Solar\textsuperscript{a} & 1 hour & 137 & 7009 & 24 & 24 & (1, 24, 168) & 7\\
           Electricity\textsuperscript{b} & 1 hour & 370 & 5833 & 24 & 24 & (1, 24, 168) & 7\\
           Traffic\textsuperscript{c} & 1 hour & 963 & 4001 & 24 & 24 & (1, 24, 168) & 7\\
           KDD-cup\textsuperscript{d} & 1 hour & 270 & 10872 & 48 & 48 & (1, 24, 168) & 1\\
           Wikipedia\textsuperscript{e} & 1 day & 2000 & 792 & 30 & 30 & (1, 7, 14)& 5\\ 
           Taxi\textsuperscript{f} & 30 mins & 1214 & 1488 & 24 & 24 & (1, 4, 12, 24, 48) & 56\\ \bottomrule
    \end{tabular}
    \vspace{1mm}
    \begin{minipage}{\linewidth}
    \footnotesize
    \textsuperscript{a}\url{https://www.nrel.gov/grid/solar-power-data.html};
    \textsuperscript{b}\url{https://archive.ics.uci.edu/dataset/321/electricityloaddiagrams20112014};
    \textsuperscript{c}\url{https://archive.ics.uci.edu/dataset/204/pems-sf};
    \textsuperscript{d}\url{https://www.kdd.org/kdd2018/kdd-cup};
    \textsuperscript{e}\url{https://github.com/mbohlkeschneider/gluon-ts/tree/mv_release/datasets};
    \textsuperscript{f}\url{https://www.nyc.gov/site/tlc/about/tlc-trip-record-data.page}. 
    \end{minipage}
    %\end{adjustbox}
\end{table}

\begin{table}[]
    \centering
    \caption{Spatiotemporal epidemic datasets used in this study.}
    \label{table:spatiotemporal_dataset_prop}

    % \scriptsize
    % \setlength{\tabcolsep}{4pt}
    % \renewcommand{\arraystretch}{1.15}

    \begin{tabular}{@{}lllccccc@{}}
    \toprule
    Country & Disease & Granularity 
    & \makecell{Locations} 
    & \makecell{Training\\Samples} 
    & \makecell{Context\\Length} 
    & \makecell{Prediction\\Length}  
    & Ref. \\
    \midrule

    Japan   & TB          & Monthly & 47 & 216 & 36 & 24 & \textsuperscript{a} \\
    USA     & ILI         & Weekly  & 50 & 328 & 26 & 13 & \textsuperscript{b} \\
    Belgium & COVID-19    & Daily   & 11 & 776 & 60 & 30 & \textsuperscript{c} \\
    Colombia & Dengue     & Weekly  & 33 & 835 & 26 & 13 & \textsuperscript{d} \\
    Hungary & Chickenpox  & Weekly  & 20 & 522 & 26 & 13 & \textsuperscript{e} \\

    \bottomrule
    \end{tabular}

    \vspace{1mm}
    \begin{minipage}{\linewidth}
    \footnotesize
    \textsuperscript{a} \cite{sumi2019time};
    \textsuperscript{b} CDC FluView 
(\url{https://gis.cdc.gov/grasp/fluview/fluportaldashboard.html});
    \textsuperscript{c} Epistat 
(\url{https://epistat.sciensano.be/covid/});
    \textsuperscript{d} \cite{clarke2024global};
    \textsuperscript{e} \cite{rozemberczki2021chickenpox}.
    \end{minipage}
\end{table}

% The Solar dataset comprises synthetic photovoltaic power generation data simulating the output of hypothetical solar plants across the United States in 2006, and is commonly used for energy forecasting studies. The Electricity dataset contains hourly electricity consumption data from 370 clients, reflecting typical power usage patterns. The Traffic dataset records road occupancy rates (ranging from 0 to 1) from 963 highway sensor stations in the San Francisco Bay Area, capturing the proportion of time road segments are occupied. The KDD-cup dataset provides multivariate air-quality observations from monitoring stations in Beijing and London, supporting short-term air-quality forecasting. The Taxi dataset provides detailed trip records from yellow, green, and for-hire vehicles in New York City, including pickup and drop-off times, locations, trip distances, fares, and passenger counts, and is widely used in urban mobility and demand forecasting studies. Finally, the Wikipedia dataset contains daily page view counts for a set of Wikipedia articles, reflecting temporal patterns in online user activity.

\subsubsection{Evaluation Metrics} We use the multivariate Continuous Ranked Probability Score ($\operatorname{CRPS}_{\text{sum}}$) for evaluating the performance of the forecasters. Additionally, we report the Normalized Root Mean Squared Error ($\operatorname{NRMSE}_{\text{sum}}$) scores achieved by Enformer and baseline models in Appendix \ref{appendix_results}. For both metrics, lower values indicate superior predictive performance. We also assess the statistical calibration of the prediction intervals generated by Enformer using Probability Integral Transform (PIT) Quantile-Quantile (Q-Q) plots. 

The CRPS is a strictly proper scoring rule \citep{gneiting2007strictly} evaluating the compatibility of a predictive cumulative distribution function $\mathcal{F}$ with an observation $u \in \mathbb{R}$:
$
\operatorname{CRPS}(\mathcal{F}, u)=\int_{\mathbb{R}}(\mathcal{F}(y)-\mathbf{1}(u \leq y))^2 d y,
$
where $\mathbf{1}$ is the indicator function. For multivariate time series, we compute $\mathrm{CRPS}_{\text{sum}}$ on the aggregated dimensions:
$$
\operatorname{CRPS}_{\text{sum}}=\mathbb{E}_t\left[\operatorname{CRPS}\left(\mathcal{F}_*, \sum_i u_t^i\right)\right],
$$
where $\mathcal{F}_*$ is the empirical distribution of the predictions summed across all dimensions. For deterministic evaluation, we utilize the NRMSE, which is defined as:
$$
\mathrm{NRMSE}=\sqrt{\frac{\operatorname{Mean}((\widetilde{Y}-Y)^2)}{\operatorname{Mean}(|Y|)}},
$$
where $\widetilde{Y}$ and $Y$ are the predicted and true series. Its multivariate extension, $\operatorname{NRMSE}_{\text{sum}}$, is similarly computed by aggregating samples across dimensions prior to evaluation.

\subsubsection{Benchmark Forecasting Models} 
To evaluate the effectiveness of the proposed models, we compare them against a range of state-of-the-art multivariate and spatiotemporal forecasting models \citep{fan2024mgtsd}. Baselines for comparing Enformer include RNN-based architectures such as Vec-LSTM, GP-scaling, GP-Copula \citep{salinas2019high}, and LSTM-MAF \citep{rasul2021multivariate}, diffusion models such as TimeGrad \citep{rasul2021autoregressive}, MG-Input \citep{fan2024mgtsd}, and NsDiff \citep{yenon}, as well as Transformer-based frameworks including Transformer-MAF \citep{rasul2021multivariate}, TACTiS \citep{drouin2022tactis}, and HDT \citep{shibo2025hdt}. Furthermore, we compare GEnformer with well-established temporal and spatiotemporal models (both point and probabilistic), as detailed in Table \ref{table:Gentransformer_Results}.

\subsubsection{Implementation Details} 
To ensure methodological consistency and fair comparison with prior studies \citep{fan2024mgtsd}, we adopt the standardized training and test splits for the temporal \citep{alexandrov2020gluonts} and spatiotemporal datasets. 

For the time series datasets, model performance is evaluated using a rolling-window framework on the test data. The number of rolling windows varies by dataset: 7 windows for the Solar, Electricity, and Traffic datasets, 5 for Wikipedia, and 1 and 56 for the KDD-cup and Taxi datasets, respectively. Before training, the datasets were standardized using the mean and standard deviation computed for each node (column). Missing values in the training portion of the Electricity dataset were handled through seasonal mean imputation. Additionally, missing observations in the test split of the KDD-cup dataset were replaced with zeros, following the procedure in \cite{fan2024mgtsd}. Similar to prior works \citep{rasul2021autoregressive, tang2021probabilistic}, we incorporate sequential lagged inputs (see Table \ref{tab:temporal_data_characteristics}) along with fixed temporal covariates, such as day-of-week and hour-of-day embeddings. The model is trained for 30 epochs using the Adam optimizer. The learning rate and batch size are selected via temporal cross-validation. 

For the spatiotemporal datasets, the context and prediction lengths are given in Table \ref{table:spatiotemporal_dataset_prop}, and the proposed model and baselines were trained for 100 epochs. While existing literature typically constrains the training ensemble size to $2$ \citep{kraft2026modeling}, our methodology introduces the number of in-sample forecast samples, $M$, as a tunable hyperparameter to better optimize model performance. During evaluation, we generate 100 out-of-sample trajectories from the trained (G)Enformer to form the forecast ensemble. The models were trained on a single NVIDIA Tesla P100 GPU with 16 GB memory. Further details regarding the hyperparameter settings and search space are provided in Appendix~\ref{app:hypers}.

\subsection{Experimental Results}

\begin{table}[h]
    \centering
    \caption{Forecasting performance on the temporal datasets in terms of the $\operatorname{CRPS}_{\text{sum}}$ metric (lower is better). Mean and (standard error) are computed from 10 independent evaluation runs. The best results are \textbf{highlighted}.}
    \label{table_crps_metric}

    \begin{tabular}{@{}lcccccc@{}}
    \toprule
    Method & Solar & Electricity & Traffic & KDD-cup & Taxi & Wikipedia \\
    \midrule

    Vec-LSTM 
    & \meanstd{0.4825}{0.0027}
    & \meanstd{0.0949}{0.0175}
    & \meanstd{0.0915}{0.0197}
    & \meanstd{0.3560}{0.1667}
    & \meanstd{0.4794}{0.0343}
    & \meanstd{0.1254}{0.0174} \\
    \addlinespace[2pt]

    GP-Scaling 
    & \meanstd{0.3802}{0.0052}
    & \meanstd{0.0499}{0.0031}
    & \meanstd{0.0753}{0.0152}
    & \meanstd{0.2983}{0.0448}
    & \meanstd{0.2265}{0.0210}
    & \meanstd{0.1351}{0.0612} \\
    \addlinespace[2pt]

    GP-Copula 
    & \meanstd{0.3612}{0.0035}
    & \meanstd{0.0287}{0.0005}
    & \meanstd{0.0618}{0.0018}
    & \meanstd{0.3157}{0.0462}
    & \meanstd{0.1894}{0.0087}
    & \meanstd{0.0669}{0.0009} \\
    \addlinespace[2pt]

    LSTM-MAF 
    & \meanstd{0.3427}{0.0082}
    & \meanstd{0.0312}{0.0046}
    & \meanstd{0.0526}{0.0021}
    & \meanstd{0.2919}{0.1486}
    & \meanstd{0.2295}{0.0082}
    & \meanstd{0.0763}{0.0051} \\
    \addlinespace[2pt]

    Transformer-MAF 
    & \meanstd{0.3532}{0.0053}
    & \meanstd{0.0272}{0.0017}
    & \meanstd{0.0499}{0.0011}
    & \meanstd{0.2951}{0.0504}
    & \meanstd{0.1531}{0.0038}
    & \meanstd{0.0644}{0.0037} \\
    \addlinespace[2pt]

    TimeGrad 
    & \meanstd{0.3335}{0.0653}
    & \meanstd{0.0232}{0.0035}
    & \bestmeanstd{0.0414}{0.0112}
    & \meanstd{0.2902}{0.2178}
    & \meanstd{0.1255}{0.0207}
    & \bestmeanstd{0.0555}{0.0088} \\
    \addlinespace[2pt]

    TACTiS 
    & \meanstd{0.4209}{0.0330}
    & \meanstd{0.0259}{0.0019}
    & \meanstd{0.1093}{0.0076}
    & \meanstd{0.5406}{0.1584}
    & \meanstd{0.2070}{0.0159}
    & -- \\
    \addlinespace[2pt]

    MG-Input 
    & \meanstd{0.3239}{0.0427}
    & \meanstd{0.0238}{0.0035}
    & \meanstd{0.0658}{0.0065}
    & \meanstd{0.2977}{0.1163}
    & \meanstd{0.1592}{0.0087}
    & \meanstd{0.0567}{0.0091} \\
    \addlinespace[2pt]

    NsDiff 
    & \meanstd{0.8825}{0.0063}
    & \meanstd{0.0249}{0.0001}
    & \meanstd{0.3143}{0.0003}
    & \meanstd{0.4791}{0.0009}
    & \meanstd{0.2346}{0.0001}
    & \meanstd{0.0636}{0.0004} \\
    \addlinespace[2pt]

    HDT
    & \meanstd{0.5555}{0.0099}
    & \meanstd{0.2217}{0.0008}
    & \meanstd{0.3318}{0.0034}
    & \meanstd{0.2768}{0.0021}
    & \meanstd{0.2830}{0.0007}
    & \meanstd{0.0856}{0.0007} \\
    \addlinespace[2pt]

    \textbf{Enformer}
    & \bestmeanstd{0.2421}{0.0027}
    & \bestmeanstd{0.0216}{0.0001}
    & \meanstd{0.0644}{0.0001}
    & \bestmeanstd{0.2468}{0.0007}
    & \bestmeanstd{0.1190}{0.0003}
    & \meanstd{0.0651}{0.0002} \\

    \bottomrule
    \end{tabular}
\end{table}

\begin{table}[]
    \centering
    \caption{Forecasting performance in terms of CRPS on the spatiotemporal epidemic datasets. Mean and standard error are computed from 50 independent evaluation runs. The best results are \textbf{highlighted}.}
    \begin{tabular}{lccccc}
    \toprule
        Method &  TB & ILI &  COVID-19 &  Dengue & Chickenpox \\ \midrule
        NHiTS & $6.72$ & $102.91$ & $165.5$ & $27.31$ & $16.25$ \\ 
        Transformers & $475.54$ & $104.94$ & $195.33$ & $22.97$ & $20.15$ \\ 
        TCN & $578.56$ & $146.24$ & $457.57$ & $44.02$ & $24.95$ \\ 
        GSTAR & $6.58$ & $179.34$ & $190.85$ & $19.15$ & $20.9$ \\ 
        STGCN & $14.54$ & $115.36$ & $374.97$ & $39.27$ & $23.85$ \\ 
        DeepAR & $6.97_{\pm 0.07}$ & $151.28_{\pm 1.97}$ & $112.08_{\pm 9.24}$ & $18.49_{\pm 0.19}$ & $12.22_{\pm 0.24}$ \\ 
        Prob-iTransformer & $11.15_{\pm 0.05}$ & $179.31_{\pm 0.44}$ & $77.24_{\pm 0.33}$ & $18.91_{\pm 0.12}$ & $\mathbf{10.05_{\pm 0.06}}$ \\ 
        GpGp & $70.39_{\pm 6.79}$ & $245.18_{\pm 1.61}$ & $546.12_{\pm 23.61}$ & $38.18_{\pm 0.29}$ & $15.23_{\pm 0.16}$ \\ 
        DiffSTG & $32.04_{\pm 0.13}$ & $193.91_{\pm 1.30}$ & $243.25_{\pm 2.53}$ & $20.65_{\pm 0.18}$ & $18.67_{\pm 0.22}$ \\ 
        \textbf{Enformer} & $6.16_{\pm 0.11}$ & ${76.66_{\pm 0.74}}$ & ${80.40_{\pm 1.17}}$ & ${19.56_{\pm 0.29}}$ & $15.17_{\pm 0.91}$ \\ 
        \textbf{GEnformer} & $\mathbf{5.26_{\pm 0.03}}$ & $\mathbf{75.25_{\pm 0.42}}$ & $\mathbf{76.93_{\pm 1.05}}$ & $\mathbf{17.1_{\pm 0.36}}$ & $13.21_{\pm 0.41}$ \\ \bottomrule
    \end{tabular}
    \label{table:Gentransformer_Results}
\end{table}

\begin{figure}
\centering
\includegraphics[width=0.9\textwidth]{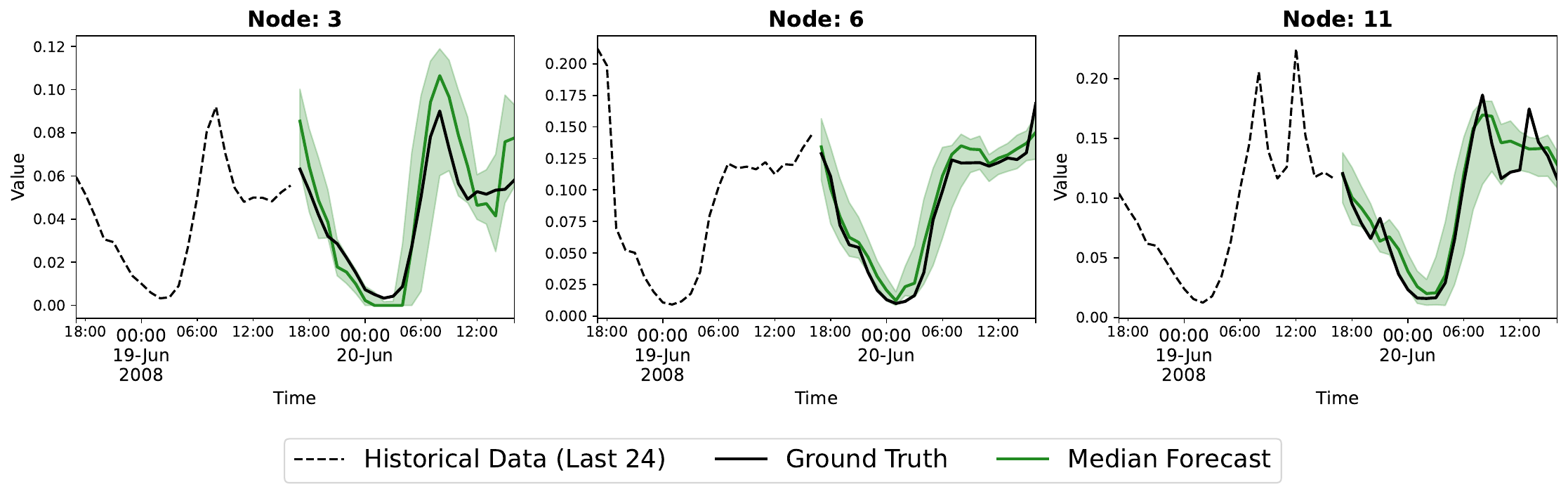}
\caption{Forecasts produced by Enformer on selected nodes of Traffic dataset on test window 5.} 
\label{fig:traffic_3}
\end{figure}

The forecasting performance of the proposed Enformer and other state-of-the-art probabilistic models on the temporal datasets is summarized in Table \ref{table_crps_metric} using the $\operatorname{CRPS}_{\text{sum}}$ metric. The table reports the mean and standard error computed from 10 independent evaluation runs with different random seeds. For the spatiotemporal epidemic datasets, the results for CRPS obtained by GEnformer across 50 independent runs and the baselines are detailed in Table \ref{table:Gentransformer_Results}. The results show that the proposed Transformer-based probabilistic frameworks achieve strong performance across multiple datasets and outperform existing baselines on several benchmarks. In particular, Enformer generates the best out-of-sample forecasts for the Solar, Electricity, KDD-cup, and Taxi datasets, significantly reducing the $\operatorname{CRPS}_{\text{sum}}$ compared to competing approaches. The lower average metrics, coupled with small standard errors, indicate that the Enformer model exhibits stable and consistent performance across repeated evaluation runs. For spatiotemporal forecasting, GEnformer outperforms the baselines on four out of five epidemic datasets. Specifically, it provides a significant performance enhancement against conventional Transformers, iTransformer, STGCN, and diffusion-based DiffSTG. Overall, the results indicate that the proposed framework can successfully capture intricate (spatio)-temporal relationships and underlying probabilistic structures in multivariate time series. Fig. \ref{fig:traffic_3} presents the out-of-sample forecasts along with 95\% prediction intervals on test window 5 of the Traffic dataset. 

%Forecasts for the first 16 nodes of each dataset are illustrated in Figs. \ref{fig:solar_forecasts}-\ref{fig:wikipedia_forecasts} in Appendix \ref{appendix:figures}. %Overall, these improvements highlight the effectiveness and robustness of the Enformer architecture for multivariate probabilistic forecasting.

\begin{figure}
\centering
\includegraphics[width=0.9\textwidth]{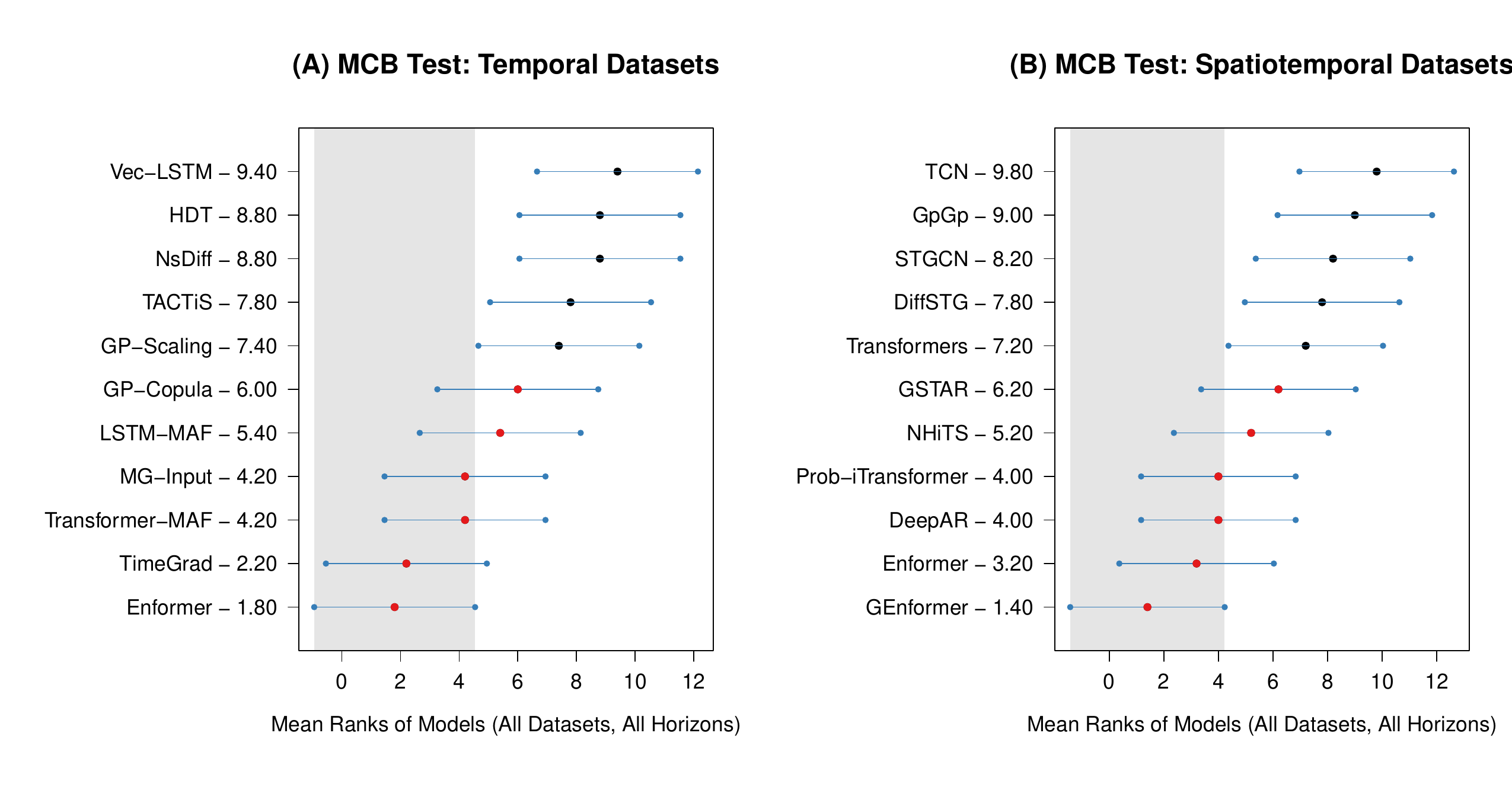}
\caption{MCB test results on the (A) temporal datasets based on $\operatorname{CRPS}_{\text{sum}}$ and (B) spatiotemporal datasets based on CRPS metric. On the y-axis, the notation $\nu$-$r$ indicates that model $\nu$ achieved an average rank of $r$.} %The gray-shaded area represents the region of statistical non-significance at $\alpha=0.05$ level.} 
\label{fig:MCB_results}
\end{figure}

Furthermore, to demonstrate the statistical robustness of the proposed model performance, we employ the Multiple Comparison with the Best (MCB) test procedure \citep{koning2005m3}. This distribution-free test ranks the forecasting frameworks based on their performance across different datasets and computes the average ranks along with their critical distance. The results of the MCB test, as depicted in Fig. \ref{fig:MCB_results}, indicate that the (A) Enformer and (B) GEnformer frameworks achieve the `best' ranks among their respective competitors based on the CRPS metric. The superior performance of GEnformer over its temporal counterpart explicitly highlights the importance of the spatial module. 

For multivariate temporal forecasting, while TimeGrad and Transformer-MAF demonstrate competitive predictive performance, the proposed Enformer offers a significant computational advantage. For instance, Enformer requires only 5.38 seconds per training epoch on the Traffic dataset, achieving a 36.9\% reduction in training time compared to Transformer-MAF and an 82.1\% reduction relative to the highly resource-intensive TimeGrad framework. Table~\ref{table:training_time} presents the average training time per epoch and average inference time for Enformer, along with two highly competitive baselines - TimeGrad and Transformer-MAF for the Traffic dataset under a similar setup \citep{tang2021probabilistic}. This demonstrates that Enformer is significantly more resource-efficient than current state-of-the-art probabilistic forecasting models. Similarly, GEnformer is lightweight compared to STGCN and DiffSTG, both of which are known for their high computational overhead.

% Additionally, the mean and standard deviation of the overall training and inference times (in seconds) for Enformer across all datasets and 10 independent runs are summarized in Table~\ref{table:all_training_times}.

\begin{table}
    \centering
    \caption{Average training time per epoch and average inference time (in seconds) for Transformer-MAF, TimeGrad, and Enformer (proposed) on the Traffic dataset.}
    \label{table:training_time}
    \begin{tabular}{ccc}
    \toprule
        Model & Training  & Testing \\ \midrule
        Transformer-MAF & $8.524\pm0.001$ & $17.835\pm0.002$ \\ 
        TimeGrad & $30.128\pm0.002$ & $44.171\pm0.003$ \\ 
        \textbf{Enformer} (Proposed) & $\mathbf{5.380 \pm 0.001}$ & $\mathbf{5.560 \pm 0.000}$ \\ \bottomrule
    \end{tabular}
\end{table}

\begin{figure}
\centering
\includegraphics[width=0.8\textwidth]{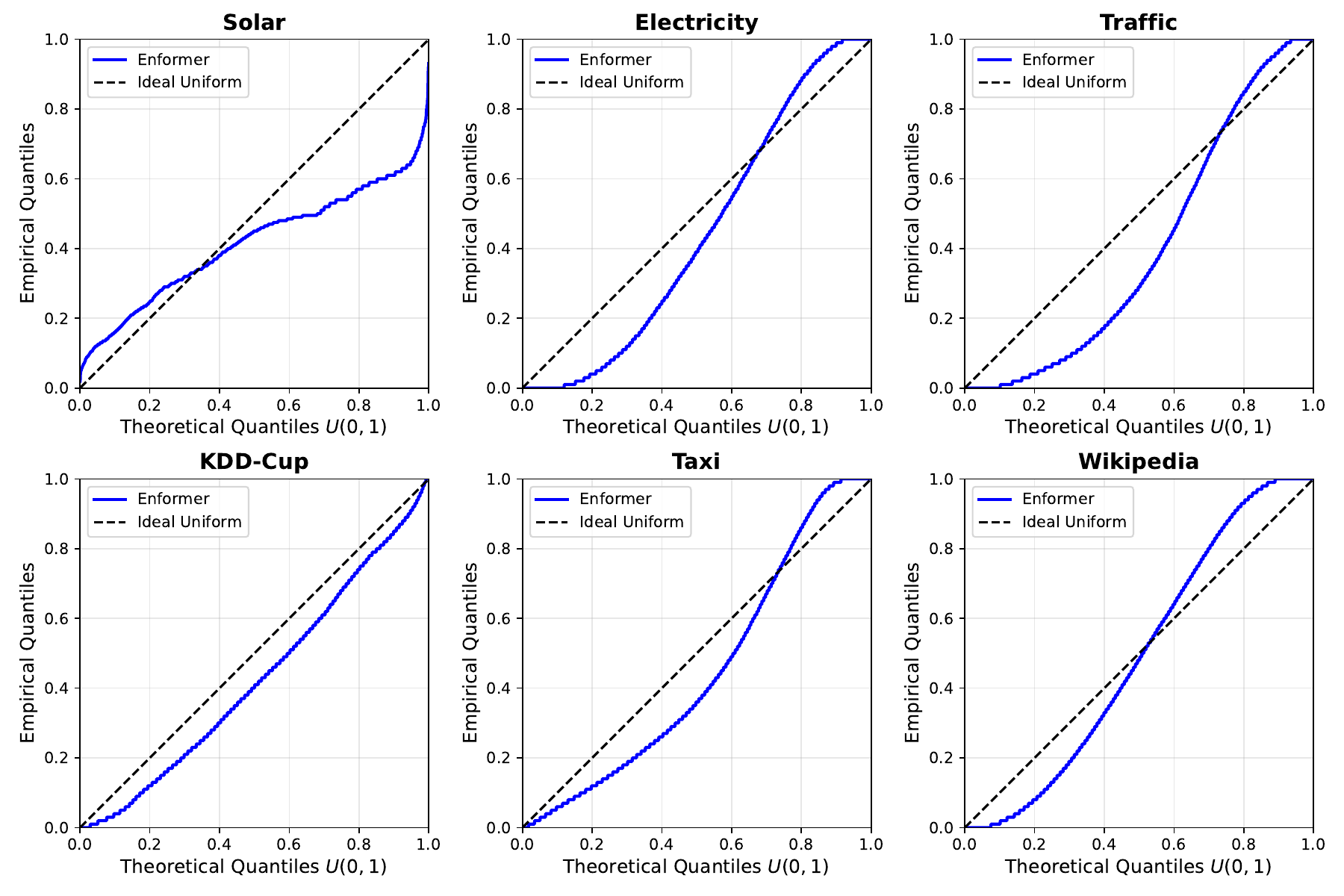}
\caption{Probability Integral Transform (PIT) Q-Q plots evaluating the predictive calibration of Enformer across the six multivariate time series datasets. The empirical quantiles (solid blue) are plotted against the theoretical quantiles of an ideal uniform distribution, $\mathcal{U}(0,1)$ (dashed black). %The test window taken for each dataset corresponds to Figs. \ref{fig:solar_forecasts}-\ref{fig:wikipedia_forecasts} in Appendix \ref{appendix:figures}.
} 
\label{fig:PIT}
\end{figure}

The forecast calibration of Enformer on the test sets of the temporal datasets is qualitatively evaluated using PIT Q-Q plots. Perfectly calibrated forecasts yield PIT values distributed as $\operatorname{Uniform}(0,1)$, corresponding to the $y=x$ diagonal. As shown in Fig. \ref{fig:PIT}, Enformer produces well-calibrated predictive distributions for the Electricity, KDD-cup, Taxi, and Wikipedia datasets, tightly aligning with the ideal diagonal. While the Solar and Traffic datasets exhibit minor deviations from the ideal uniform line, the model's overall probabilistic reliability remains robust across domains.

\subsection{Sensitivity Analysis}
To better understand the internal dynamics of the proposed Enformer, we conducted a sensitivity analysis to analyze the impact of the in-sample training ensemble size ($M$). We evaluated this hyperparameter across all six temporal datasets, tracking both the computational overhead and the resulting probabilistic forecasting accuracy.

\begin{figure}[]
\centering
\includegraphics[width=0.8\textwidth]{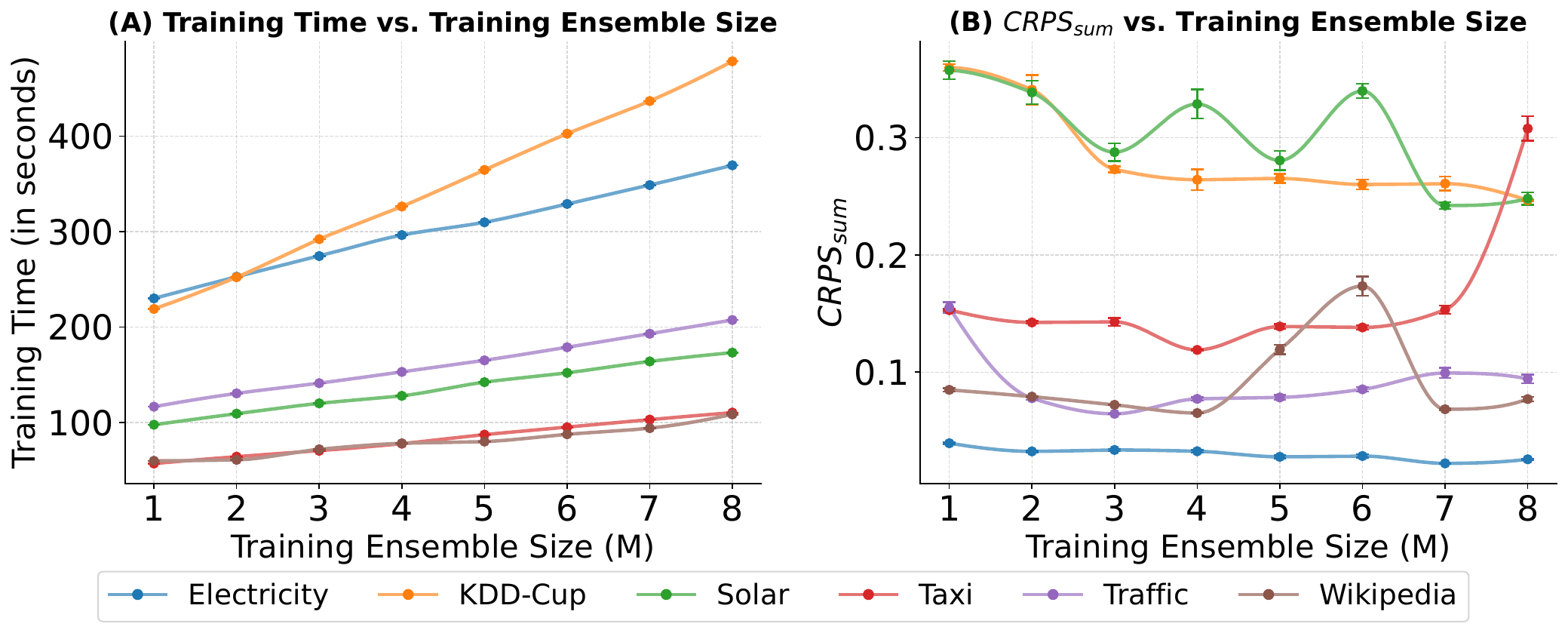}
\caption{Sensitivity analysis studying the effect of the in-sample training ensemble size ($M$) on the proposed Enformer. (A) Computational overhead represented by the total training time in seconds; (B) Forecasting performance evaluated by the $\operatorname{CRPS}_{\text{sum}}$ metric across the six datasets. The circular points represent the mean values and the vertical error lines indicate the standard deviation computed over 10 independent runs.} 
\label{fig:ablation_study}
\end{figure}

%\subsubsection{Computational Cost.} 
Fig. \ref{fig:ablation_study} (A) illustrates the relationship between the training ensemble size ($M$) and the total training time (in seconds). As $M$ scales from 1 to 8, the computational cost increases linearly across all datasets. While the base training time and the slope of this increase vary depending on the inherent size and complexity of each dataset (ranging from under 100 seconds to over 400 seconds), the linear scaling factor remains consistent. Notably, the KDD-cup dataset exhibits the steepest slope in the training time versus $M$ curve, as it contains the largest number of training examples among all the evaluated datasets. This confirms that generating more in-sample forecast trajectories linearly bottlenecks the training phase, complementing the complexity analysis in Sec. \ref{sec:complexity_analysis}.

The relationship between $M$ and the model's predictive accuracy, measured by the $\operatorname{CRPS}_{\text{sum}}$ metric, is highly non-linear. Fig. \ref{fig:ablation_study} (B) demonstrates how the $\operatorname{CRPS}_{\text{sum}}$ scores respond as the training ensemble size increases from 1 to 8. For most datasets, the forecasting error either stabilizes or reaches a minimum at lower-to-mid values of $M$, specifically between $M=2$ and $M=5$. Pushing the ensemble size beyond this optimal window to $M=7$ or $M=8$ does not consistently improve performance. As the model's performance depends on $M$, we optimize $ M$ in the main experiments rather than setting it to $2$ as in prior works.

\section{Conclusion} \label{sec:conclusion}

In this paper, we introduced Enformer and GEnformer, which are frameworks for multivariate probabilistic forecasting. The proposed methods integrate the engression paradigm with Transformer architectures. By combining attention-based sequence modeling with noise-driven generative learning, the proposed approaches enable the learning of conditional predictive distributions while remaining computationally lightweight. This allows the models to generate diverse forecast trajectories and provide uncertainty-aware predictions without relying on restrictive parametric assumptions. Experimental results on benchmark datasets demonstrate that (G)Enformer achieves competitive forecasting accuracy while producing well-calibrated probabilistic forecasts. These results highlight the effectiveness of combining attention mechanisms with generative uncertainty modeling for multivariate time series forecasting.

Future work may explore extensions that integrate domain-specific constraints to further improve probabilistic forecasts for spatiotemporal data. Furthermore, while the proposed framework demonstrates strong intrinsic calibration, any domain-specific deviations can be addressed by applying established post-hoc recalibration techniques to the output ensembles. Such standard post-processing methods efficiently refine the predictive distributions without requiring architectural modifications.

\section*{Acknowledgement}
The authors thank Ms. Donia Besher for her help with implementing the GEnformer model.

\section*{Data and Code Availability}
All datasets used in this study are publicly available, and the sources are listed in Tables \ref{tab:temporal_data_characteristics} and \ref{table:spatiotemporal_dataset_prop}. Code is available at \url{{https://github.com/yuvrajiro/genformer}}, as well as through our Python package \href{https://pypi.org/project/genformer/}{\texttt{genformer}}\footnote{Documentation: \url{https://genformer.readthedocs.io/}}.

% \begin{credits}
% \subsubsection{\ackname} A bold run-in heading in small font size at the end of the paper is
% used for general acknowledgments, for example: This study was funded

% \subsubsection{\discintname}
% The authors report no competing interests for this article.
% \end{credits}

\bibliography{sn-bibliography}

% \bibliography{Bibliography}
% \clearpage
% \section*{Appendix}
\section*{Appendix}
\appendix

\section{Model Hyperparameters}\label{app:hypers}
In the proposed architectures, we optimize the model hyperparameters via a temporal validation approach using the Optuna framework \citep{akiba2019optuna}. In Enformer, the number of attention heads ($n_{\text{head}}$) is selected from the set $\{2,4,6,\ldots,16\}$. The model dimension, which determines the feature size of the inputs to both the Transformer encoder and decoder, is defined as $ d_{\text{model}} = n_{\text{head}} \times d_{\text{model multiplier}},$ where $d_{\text{model multiplier}} \in \{8,9,10,\ldots,24\}$ is an integer hyperparameter. This ensures that $d_{\text{model}}$ is always divisible by the number of attention heads. The number of encoder and decoder layers is jointly selected from the set $\{1,2,3,4\}$, with the same value used for both components. The dimension of the feedforward network is chosen from $\{128,256,512,1024\}$. The dropout rate is sampled from the interval [0,0.4]. We consider multiple activation functions, including GLU, Bilinear, ReGLU, GEGLU, SwiGLU, ReLU, and GELU. The learning rate is selected from the log-uniform range $\left[10^{-5}, 10^{-2}\right]$, while the batch size is chosen from $\{32, 64, 128\}$. Additionally, the noise distribution type \{Uniform, Gaussian\}, noise standard deviation within $[0.5,2.0]$, and the training ensemble size $M$ from $\{2,3, \ldots, 8\}$ are tuned. For GEnformer, the embedding dimension of the GCN layers, was chosen between $[2, 64]$. The final reported results correspond to the best-performing configurations obtained from the hyperparameter searches. 

\section{Forecasting Performance}\label{appendix_results}
We compute $\operatorname{NRMSE}_{\text{sum}}$ on the median of the forecast ensembles, which constitutes the point forecast of our model, and report the results in Table~\ref{table_nrmse_metric}. We also include the performance of the Autoformer \citep{wu2021autoformer}, PatchTST \citep{nie2022time}, $\mathrm{D}^3$ VAE \citep{li2022generative}, and TimeDiff \citep{shen2023non} frameworks from \cite{fan2024mgtsd}. The results indicate that the proposed Enformer model outperforms all baselines across all datasets except Traffic, where it is the second-best.

\begin{table}[]
    \centering
    \footnotesize
    \caption{Forecasting performance on the temporal datasets in terms of the $\operatorname{NRMSE}_{\text{sum}}$ metric (lower is better). Best results are \textbf{highlighted}.}
    \label{table_nrmse_metric}
    
    \begin{tabular}{@{}lcccccc@{}}
    \toprule
    Method & Solar & Electricity & Traffic & KDD-cup & Taxi & Wikipedia \\
    \midrule

    Vec-LSTM
    & \meanstd{0.9952}{0.0077}
    & \meanstd{0.1439}{0.0228}
    & \meanstd{0.1451}{0.0248}
    & \meanstd{0.4461}{0.1833}
    & \meanstd{0.6398}{0.0390}
    & \meanstd{0.1618}{0.0162} \\
    \addlinespace[2pt]

    GP-Scaling
    & \meanstd{0.9004}{0.0095}
    & \meanstd{0.0811}{0.0062}
    & \meanstd{0.1469}{0.0181}
    & \meanstd{0.3445}{0.0621}
    & \meanstd{0.3598}{0.0285}
    & \meanstd{0.1710}{0.1006} \\
    \addlinespace[2pt]

    GP-Copula
    & \meanstd{0.8279}{0.0053}
    & \meanstd{0.0512}{0.0009}
    & \meanstd{0.1282}{0.0033}
    & \meanstd{0.2605}{0.0227}
    & \meanstd{0.3125}{0.0113}
    & \meanstd{0.0930}{0.0076} \\
    \addlinespace[2pt]

    Autoformer 
    & \meanstd{0.7046}{0.0000}
    & \meanstd{0.0475}{0.0000}
    & \meanstd{0.0951}{0.0000}
    & \meanstd{0.8984}{0.0000}
    & \meanstd{0.3498}{0.0000}
    & \meanstd{0.1052}{0.0000} \\
    \addlinespace[2pt]

    PatchTST
    & \meanstd{0.7270}{0.0000}
    & \meanstd{0.0474}{0.0000}
    & \meanstd{0.1897}{0.0000}
    & \meanstd{0.5137}{0.0000}
    & \meanstd{0.3690}{0.0000}
    & \meanstd{0.0915}{0.0000} \\
    \addlinespace[2pt]

    $\mathrm{D}^3$ VAE
    & \meanstd{0.7472}{0.0508}
    & \meanstd{0.1640}{0.0928}
    & \meanstd{0.4722}{0.1197}
    & \meanstd{0.5628}{0.0419}
    & \meanstd{0.7624}{0.5598}
    & \meanstd{2.2094}{2.1646} \\
    \addlinespace[2pt]

    TimeDiff
    & \meanstd{1.5985}{0.0359}
    & \meanstd{0.3714}{0.0073}
    & \meanstd{0.5520}{0.0087}
    & \meanstd{0.4955}{0.0147}
    & \meanstd{0.5479}{0.0084}
    & \meanstd{0.1412}{0.0099} \\
    \addlinespace[2pt]

    TimeGrad
    & \meanstd{0.6953}{0.0845}
    & \meanstd{0.0348}{0.0057}
    & \bestmeanstd{0.0653}{0.0244}
    & \meanstd{0.4092}{0.1332}
    & \meanstd{0.2365}{0.0386}
    & \meanstd{0.0870}{0.0106} \\
    \addlinespace[2pt]

    TACTiS
    & \meanstd{0.8532}{0.0851}
    & \meanstd{0.0427}{0.0023}
    & \meanstd{0.2270}{0.0159}
    & \meanstd{0.6513}{0.1767}
    & \meanstd{0.3387}{0.0097}
    & -- \\
    \addlinespace[2pt]

    NsDiff
    & \meanstd{1.1191}{0.0054}
    & \meanstd{0.0417}{0.0002}
    & \meanstd{0.4101}{0.0001}
    & \meanstd{0.6324}{0.0005}
    & \meanstd{0.3176}{0.0001}
    & \meanstd{0.1018}{0.0005} \\
    \addlinespace[2pt]

    HDT
    & \meanstd{1.2150}{0.0144}
    & \meanstd{0.3045}{0.0006}
    & \meanstd{0.5439}{0.0039}
    & \meanstd{0.3925}{0.0031}
    & \meanstd{0.4528}{0.0013}
    & \meanstd{0.1389}{0.0007} \\
    \addlinespace[2pt]

    % MG-TSD
    % & \meanstd{0.6178}{0.0418}
    % & \meanstd{0.0241}{0.0030}
    % & \meanstd{0.0563}{0.0230}
    % & \meanstd{0.3001}{0.0997}
    % & \meanstd{0.2334}{0.0313}
    % & \meanstd{0.0810}{0.0057} \\
    % \addlinespace[2pt]

    \textbf{Enformer}
    & \bestmeanstd{0.3129}{0.0095}
    & \bestmeanstd{0.0278}{0.0004}
    & \meanstd{0.0789}{0.0015}
    & \bestmeanstd{0.3094}{0.0086}
    & \bestmeanstd{0.1516}{0.0014}
    & \bestmeanstd{0.0830}{0.0005} \\

    \bottomrule
    \end{tabular}
\end{table}

\end{document}